\documentclass{article}

\usepackage[final]{neurips_2020}

\usepackage{graphicx}
\usepackage{caption}
\usepackage{xcolor}
\usepackage{colortbl}
\usepackage{subcaption}
\usepackage{float}
\usepackage{verbatimbox}
\usepackage{algorithm}
\usepackage[noend]{algpseudocode}

\algrenewcommand\algorithmicforall{\textbf{foreach}}
\algrenewcommand\algorithmicindent{.8em}
\usepackage{booktabs}
\usepackage{multirow}
\usepackage{natbib}
\usepackage{listings}
\usepackage[linguistics]{forest}
\usepackage{dsfont}
\usepackage{caption}
\usepackage{amsmath}
\usepackage[utf8]{inputenc} 
\usepackage[T1]{fontenc}    
\usepackage{hyperref}       
\usepackage{url}            
\usepackage{booktabs}       
\usepackage{amsfonts}       
\usepackage{nicefrac}       
\usepackage{microtype}      
\RequirePackage[textsize=tiny,backgroundcolor=orange!20,linecolor=red!50]{todonotes}%
\DeclareMathOperator*{\argmax}{argmax}
\title{BOSS: Bayesian Optimization over String Spaces}

\author{%
  Henry B. Moss \\
  STOR-i Centre for Doctoral Training\\
  Lancaster University, UK\\
  \texttt{h.moss@lancaster.ac.uk} \\
   \And
   Daniel Beck \\
   Computing and Information Systems \\
   University of Melbourne, Australia \\
   \texttt{d.beck@unimelb.edu.au} \\
   \AND
   Javier Gonz{\'a}lez \\
   Microsoft Research  \\
   Cambridge, UK\\
   \And
   David S. Leslie \\
   Dept. of Mathematics \\and Statistics \\
   Lancaster University, UK \\
   \And
   Paul Rayson \\
   School of Computing \\and Communications \\
   Lancaster University, UK \\
}

\begin{document}

\maketitle

\begin{abstract}

This article develops a Bayesian optimization (BO) method which acts directly over raw strings, proposing the first uses of string kernels and genetic algorithms within BO loops. Recent applications of BO over strings have been hindered by the need to map inputs into a smooth and unconstrained latent space. Learning this projection is computationally and data-intensive. Our approach instead builds a powerful Gaussian process surrogate model based on string kernels, naturally supporting variable length inputs, and performs efficient acquisition function maximization for spaces with syntactical constraints. Experiments demonstrate considerably improved optimization over existing approaches across a broad range of constraints, including the popular setting where syntax is governed by a context-free grammar.

\end{abstract}

\section{Introduction}

Many tasks in chemistry, biology and machine learning can be framed as optimization problems over spaces of strings. Examples include the design of synthetic genes \citep{gonzalez2015bayesian,tanaka2018bayesian} and chemical molecules \citep{griffiths2017constrained,gomez2018automatic}, as well as problems in symbolic regression \citep{kusner2017grammar} and  kernel design \citep{lu2018structured}. Common to these applications is the high cost of evaluating a particular input, for example requiring resource and labor-consuming wet lab tests. Consequently, most standard discrete optimization routines are unsuitable, as they require many evaluations. 

\begin{figure}[t]
\centering
\begin{minipage}[b]{.32\textwidth}
  \centering
\includegraphics[width=\textwidth]{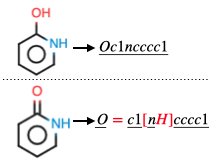}
\caption{Similar molecules have SMILES strings with local differences (red) but common non-contiguous sub-sequences.}
\label{SMILES_EXAMPLE}
\end{minipage}%
\hfill
\begin{minipage}[b]{.64\textwidth}
  \centering
  \includegraphics[width=\textwidth]{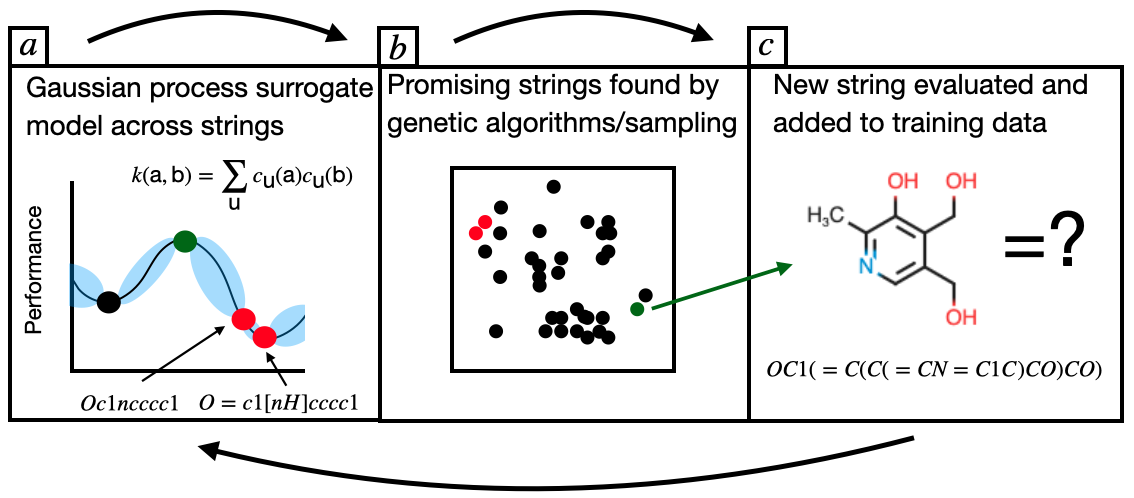}
  \captionof{figure}{BO loop for molecule design using a string kernel surrogate model (a) and genetic algorithms for acquisition function maximization (b).}
    \label{workflow}
\end{minipage}
\vskip -0.5cm
\end{figure}

Bayesian Optimization \citep[BO]{shahriari2015taking} has recently risen as an effective strategy to address the applications above, due to its ability to find good solutions within heavily restricted evaluation budgets. However, the vast majority of BO approaches assume a low dimensional, mostly continuous space; string inputs have to be converted to fixed-size vectors such as bags-of-ngrams or latent representations learned through an unsupervised model, typically a variational autoencoder \citep[VAE]{kingma2013auto}. In this work, we remove this encoding step and propose a BO architecture that operates directly on raw strings through the lens of convolution kernels \citep{haussler1999convolution}. In particular, we employ a Gaussian Process \citep[GP]{rasmussen2003gaussian} with a {\em string kernel} \citep{lodhi2002text} as the surrogate model for the objective function, measuring the similarity between strings by examining shared non-contiguous sub-sequences. String kernels provide an easy and user-friendly way to deploy BO loops directly over strings, avoiding the expensive architecture tuning required to find a useful VAE. At the same time, by using a kernel trick to work in much richer feature spaces than the bags-of-ngrams vectors, string kernels can encode the non-contiguity known to be informative when modeling genetic sequences \citep{vert:hal-00012124} and SMILES \citep{anderson1987smiles} representations of molecules \citep{cao2012silico}(see Figure \ref{SMILES_EXAMPLE}). We show that our string kernel's two parameters can be reliably fine-tuned to model complex objective functions with just a handful of function evaluations, without needing the large collections of unlabeled data required to train VAEs. 

Devising a BO framework directly over strings raises the question of how to maximize {\em acquisition functions}; heuristics used to select new evaluation points. Standard BO uses numerical methods to maximize these functions but these are not applicable when the inputs are discrete structures such as strings. To address this challenge, we employ a suite of genetic algorithms \citep{whitley1994genetic} to provide efficient exploration of string spaces under a range of syntactical constraints.

Our contributions can be summarized as follows:

\begin{itemize}
\itemsep0em 
    \item We introduce string kernels into BO, providing powerful GP surrogate models of complex objective functions with just two data-driven parameters (Figure \ref{workflow}.a).
    \item We propose a suite of genetic algorithms suitable for efficiently optimizing acquisition functions under a variety of syntactical constraints (Figure \ref{workflow}.b).
    \item We demonstrate that our framework out-performs established baselines across four scenarios encompassing a range of applications and diverse set of constraints.
\end{itemize}

\section{Related Work}
\label{sec:related}

\paragraph{BO by feature extraction}

BO has previously been applied to find genes with desirable features: a high-cost string optimization problem across a small alphabet of four bases. Genes are represented as either codon frequencies (a bags-of-ngrams of triplets of characters) \citep{gonzalez2015bayesian}, or as a one-hot-encoding of the genes at each location in the string \citep{tanaka2018bayesian}. Although these representations are sufficient to allow BO to improve over random gene designs, each mapping discards information known to be important when modeling genes. A bags-of-ngrams  representation ignores positional and contextual information by modeling characters to have equal effect regardless of position or context, whereas a one-hot encoding fails to exploit translational invariance. Moreover, by assuming that all potential genes belong to a small fixed set of candidates, \cite{gonzalez2015bayesian} and \cite{tanaka2018bayesian} ignore the need to provide an efficient acquisition optimization routine. This assumption is unrealistic for many real gene design loops and is tackled directly in our work.

\paragraph{BO with VAEs}

\cite{kusner2017grammar,gomez2018automatic} and \cite{lu2018structured} use VAEs to learn latent representations for string spaces following the syntactical constraints given by context-free grammars (CFG). Projecting a variable-length and constrained string space to an unconstrained latent space of fixed dimensions requires a sophisticated mapping, which in turn requires a lot of data to learn. As BO problems never have enough string-evaluation pairs to learn a supervised mapping, VAEs must be trained to reconstruct a large collection of valid strings sampled from the CFG. A representation learned in this purely unsupervised manner will likely be poorly-aligned with the problem's objective function, under-representing variation and over-emphasizing sub-optimal areas of the original space. Consequently, VAE's often explore only limited regions of the space and have `dead' areas that decode to invalid strings \citep{griffiths2017constrained}.  Moreover, performance is sensitive to the arbitrary choice of the closed region of latent space considered for BO.

\paragraph{Evolutionary algorithms in BO}

The closest existing idea to our work is that of \cite{kandasamy2018neural}, where an evolutionary algorithm optimizes acquisition functions over a space of neural network architectures.  However, their approach does not support syntactically constrained spaces and, as it is based solely on local mutations, cannot perform the global search required for large string spaces. Moreover, as their kernel is based on an \textit{optimal transport} distance between individual network layers, it does not model the non-contiguous features supported by string kernels. Contemporaneous work of \cite{swersky2020amortized} also considers BO over strings and proposes an evolutionary algorithm based on generative modeling for their acquisition function optimization. However, their approach relies on ensembles of neural networks rather than GP surrogate models, is suitable for strings of up to only $100$ characters and does not support spaces with syntactic constraints.

\section{Preliminaries}
\label{sec:bg}

\paragraph{Bayesian Optimization}

In its standard form, BO seeks to maximize a smooth function $f:\mathcal{X}\rightarrow\mathds{R}$ over a compact set $\mathcal{X}\subset\mathds{R}^d$ in as few evaluations as possible. Smoothness is exploited to predict the performance of not yet evaluated points, allowing evaluations to be focused into promising areas of the space. BO loops have two key components - a \textit{surrogate model} and an \textit{acquisition function}. 

\textit{Surrogate model} To predict the values of $f$ across $\mathcal{X}$,  a surrogate model is fit to the previously collected (and potentially noisy) evaluations $D_t=\{(\textbf{x}_i,y_i)\}_{i=1,..,t}$, where $y_i=f(\textbf{x}_i)+\epsilon_i$ for iid Gaussian noise $\epsilon_i\sim \mathcal{N}(0,\sigma^2)$. As is standard in the literature, we use a GP surrogate model \citep{rasmussen2003gaussian}. A GP provides non-parametric regression of a particular smoothness controlled by a kernel $k:\mathcal{X}\times\mathcal{X}\rightarrow\mathds{R}$ measuring the similarity between two points.

\textit{Acquisition function} The other crucial ingredient for BO is an acquisition function $\alpha_t
:\mathcal{X}\rightarrow \mathds{R}$, measuring the utility of making a new evaluation given the predictions of our surrogate model. We use the simple yet effective search strategy of expected improvement (EI): evaluating points yielding the largest improvement over current evaluations. Although any BO acquisition function is compatible with our framework, we choose EI as it provides an effective search whilst not incurring significant BO overheads. Under a GP, EI has a closed form expression and can be efficiently calculated (see \cite{shahriari2015taking}). A single BO loop is completed by evaluating the location with maximal utility $\textbf{x}_{t+1}=\argmax_{\textbf{x}\in\mathcal{X}}\alpha_t(\textbf{x})$ and is repeated until the optimization budget is exhausted.

\paragraph{String Kernels (SKs)}

\begin{table}[t]
\begin{tabular}{p{0.5\textwidth} p{0.5\textwidth}}
  \vspace{0pt} 
\begin{tabular}{c|ccc}
 & \multicolumn{3}{c}{Sub-sequence Occurrence, $\textbf{u}$} \\ \cline{2-4} 
String, $\textbf{s}$ & "genic" & "geno" & "ge" \\ \hline
"genetics" & "\underline{\textbf{gen}}et\underline{\textbf{ic}}s" &  & \begin{tabular}[c]{@{}c@{}}"\underline{\textbf{ge}}netics"\\ "\underline{\textbf{g}}en\underline{\textbf{e}}tics"\end{tabular} \\[0.4cm]
"genomic" & "\underline{\textbf{gen}}om\underline{\textbf{ic}}" & "\underline{\textbf{geno}}mic" & "\underline{\textbf{ge}}nomic" \\[0.2cm]
"genomes" &  & "\underline{\textbf{geno}}mes" & \begin{tabular}[c]{@{}c@{}}"\underline{\textbf{ge}}nomes"\\ "\underline{\textbf{g}}enom\underline{\textbf{e}}s"\end{tabular}
\end{tabular}
  &
  \vspace{0pt} 
  \hspace{20pt}
  \begin{tabular}{|ccc}
\multicolumn{3}{|c}{Sub-sequence Contribution,  $c_{\textbf{u}}(\textbf{s})$} \\ \hline
"genic" & "geno" & "ge" \\ \hline 
$\lambda_m^5\lambda_g^2$ & $0$ & $\lambda_m^2(1+\lambda_g^2)$ \\[0.5cm]
$\lambda_m^5\lambda_g^2$ & $\lambda_m^4$ & $\lambda_m^2$ \\[0.5cm]
$0$ & $\lambda_m^4$ & $\lambda_m^2(1+\lambda_g^4)$
\end{tabular}
\end{tabular}
\caption{Occurrences (left panel) and respective contributions function values (right panel) of sample sub-sequences when evaluating the strings "genetics", "genomic" and "genomes".}
\label{Kernel_example}
\vskip -0.5cm
\end{table}

SKs are a family of kernels that operate on strings, measuring their similarity through the number of {\em sub-strings} they share. Specific SKs are then formally defined by the particular definition of a sub-string they encompass, which defines the underlying feature space of the kernel. In this work, we employ the {\em Sub-sequence String Kernel} (SSK) \citep{lodhi2002text,cancedda2003word}, which uses {\em sub-sequences} of characters as features. The sub-sequences can be non-contiguous, giving rise to an exponentially-sized feature space. While enumerating such a space would be infeasible, the SSK uses the kernel trick to avoid computation in the primal space, enabled via an efficient dynamic programming algorithm. By matching occurrences of sub-sequences, SSKs can provide a rich contextual model of string data, moving far beyond the capabilities of popular bag-of-ngrams representations where only contiguous occurrences of sub-strings are modeled.

Formally, an $n^{th}$ order SSK between two strings $\textbf{a}$ and $\textbf{b}$ is defined as
\begin{align}
{k}_n(\textbf{a},\textbf{b})=\sum_{\textbf{u}\in \Sigma^n} c_{\textbf{u}}(\textbf{a})c_{\textbf{u}}(\textbf{b})\quad \mbox{ for } \quad c_{\textbf{u}}(\textbf{s})=\lambda_m^{|\textbf{u}|}\sum_{1<i_1<..<i_{|\textbf{u}|}<|\textbf{s}|}\lambda_g^{i_{|\textbf{u}|}-i_1} \mathds{1}_{\textbf{u}}((s_{i_1},..,s_{i_{|\textbf{u}|}})),
    \label{String_Kernel}\nonumber
\end{align}
where $\Sigma^n$ denotes the set of all possible ordered collections containing up to $n$ characters from our alphabet $\Sigma$, $\mathds{1}_{\textbf{x}}(\textbf{y})$ is the indicator function checking if the strings $\textbf{x}$ and $\textbf{y}$ match, and the match decay $\lambda_m\in[0,1]$  and gap decay $\lambda_g\in[0,1]$ are kernel hyper-parameters. Intuitively, $c_{\textbf{u}}(\textbf{s})$ measures the contribution of sub-sequence $\textbf{u}$ to string $\textbf{s}$, and the choices $\lambda_m$ and $\lambda_g$ control the relative weighting of long and/or highly non-contiguous sub-strings (Table \ref{Kernel_example}). To allow the meaningful comparison of strings of varied lengths, we use a normalized string kernel $\tilde{k}_n(\textbf{a},\textbf{b}) = {k}_n(\textbf{a},\textbf{b}) / \sqrt{\smash[b]{{k}_n(\textbf{a},\textbf{a}){k}_n(\textbf{b},\textbf{b})}}$.


\section{Bayesian Optimization Directly On Strings}
In string optimization tasks, we seek the optimizer $\textbf{s}^* = \argmax_{\textbf{s}\in S} f(\textbf{s})$ of a function $f$ across a set of strings $S$. In this work, we consider different scenarios for $S$ arising from three different types of syntactical constraints and a sampling-based approach for when constraints are not fully known. In Section \ref{experiments} we demonstrate the efficacy of our proposed framework across all four scenarios.
\begin{enumerate}
\itemsep0em 
    \item \textbf{Unconstrained} Any string made exclusively from characters in the alphabet $\Sigma$ are allowed. $S$ contains all these strings of any (or a fixed) length.
    \item \textbf{Locally constrained} $S$ is a collection of strings of fixed length, where the set of possible values for each character depends on its position in the string, i.e. the character $s_i$ at location $i$ belongs to the set $\Sigma_i\subseteq\Sigma$.
    \item \textbf{Grammar constrained} $S$ is the set of strings made from $\Sigma$ that satisfy the syntactical rules specified by a context-free grammar.
    \item \textbf{Candidate Set}. A space with unknown or very complex syntactical rules, but for which we have access to a large collection $S$ of valid strings.
\end{enumerate}

\subsection{Surrogate Models for String Spaces}
\label{subsec:string}

To build a powerful model across string spaces, we propose using an SSK within a GP. However, the vanilla SSK presented above is not immediately suitable due to its substantial computational cost. In contrast to most applications of GPs, BO surrogates are trained on small datasets and so the computational bottleneck is not inversion of Gram matrices. Instead, the primary contributors to cost are the many kernel evaluations required to maximize acquisition functions. Therefore, we develop two modifications to improve the efficiency and scalability of our SSK.

\textit{Efficiency} Using the dynamic program proposed by \cite{lodhi2002text}, obtaining a single evaluation of an $n^{th}$ order SSK is $O(nl^2)$, where $l=\max(|\textbf{a}|,|\textbf{b}|)$. For our applications where many kernel evaluations are to be made in parallel, we found the vectorized formulation of \cite{beck2017learning} to be more appropriate. Although, having a larger complexity of $O(nl^3)$, a vectorized formulation can exploit recent advancements in parallel processing and in practice was substantially faster. Moreover, \cite{beck2017learning}'s formulation provides gradients with respect to the kernel parameters, allowing their fine-grained tuning to a particular optimization task. We found the particular string kernel proposed by \cite{beck2017learning} --- with individual weights for each different sub-sequence length --- to be overly flexible for our BO applications. We adapt their recursive algorithm for our SSK (Appendix \ref{appendix:dynamicprogram}).

\textit{Scalability} Even with a vectorized implementation, SSKs are computationally demanding for long strings. Comprehensively tackling the scalability of string kernels is beyond the scope of this work and is an area of future research. However, we perform a simple novel approximation to allow demonstrations of BO for longer sequences: we split sequences into $m$ parts, applying separate string kernels (with tied kernel parameters) to each individual part and summing their values. This reduces the complexity of kernel calculations from $O(nl^3)$ to 
$O(nl^3/m^2)$
without a noticeable effect on performance (Section \ref{genetics}). Moreover, the $m$ partial kernel calculations can be computed in parallel.

\subsection{Acquisition function optimization over String Spaces}
\label{subsec:ga}

We now present a suite of routines providing efficient acquisition function optimization under different types of syntactical constraints. In particular, we propose using \textit{genetic algorithms} (GA) \citep{whitley1994genetic}, biologically inspired optimization routines that successively evaluate and evolve populations of $n$ strings. Candidate strings undergo one of two stochastic perturbations: a \textit{mutation} operation producing a new offspring string from a single parent, and a \textit{crossover} operation combining attributes of two parents to produce two new offspring. GAs are a natural choice for optimizing acquisition functions as they avoid local maxima by maintaining diversity and the evolution can be carefully constructed to ensure compliance to syntactical rules. We stress that GAs require many function evaluations and so are not suitable for optimizing a high-cost objective function, just for this `inner-loop' maximization. To highlight robustness, the parameters of our GAs are not tuned to our individual experiments (Appendix \ref{appendix:AG}). When  syntactical rules are poorly understood and cannot be encoded into the optimization, we recommend the simple but effective strategy of maximizing acquisition functions across a random sample of valid strings.

\paragraph{GAs for unconstrained and locally constrained string spaces}

For our first two types of syntactical constraints, standard definitions of crossover and mutation are sufficient. For mutation, a random position $i$ is chosen and the character at this point is re-sampled uniformly from the set of permitted characters $\Sigma_i$ (or just $\Sigma$) for locally constrained (unconstrained) spaces. For crossover, a random location is chosen within one of the parent strings and the characters up until the crossover point are swapped between the parents. Crucially, the relative positions of characters in the strings are not changed by this operation and so the offspring strings still satisfy the space's constraints. 

\paragraph{GA for grammar-constrained string spaces}

\begin{figure}[t]
\centering
\begin{minipage}{.48\textwidth}
  \centering
\includegraphics[width=0.9\textwidth]{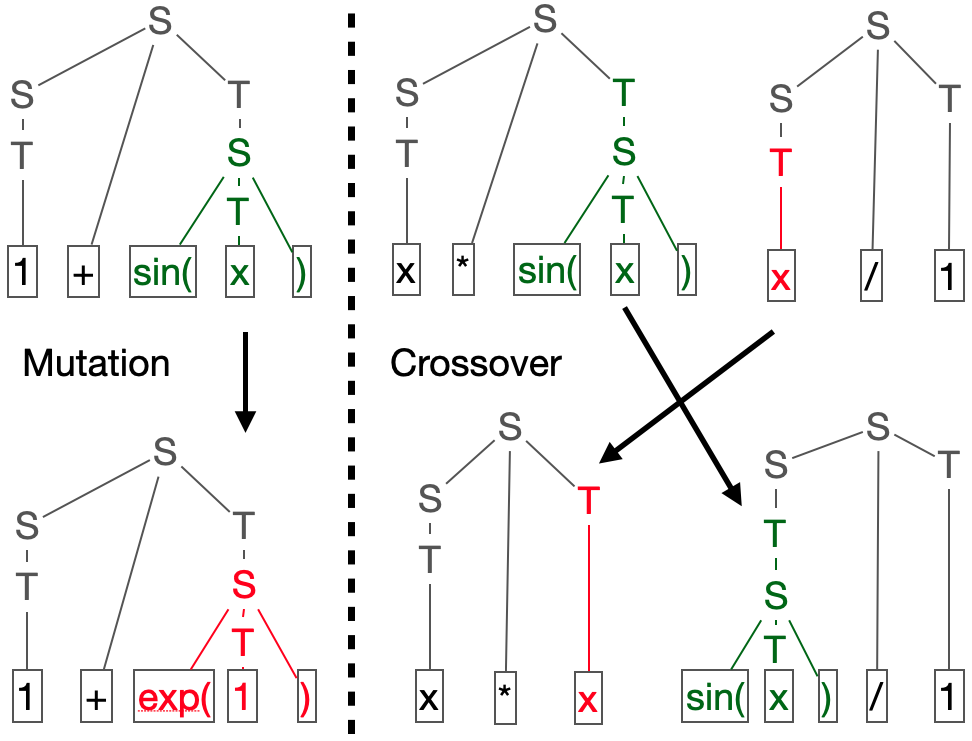}
\caption{Mutations and crossover of arithmetic expressions following the grammar:\\
 \lstinline{S -> S `+' T | S `*' T | S `/' T | T}\\
  \lstinline{T -> `sin(' S `)' | `exp(' S `)' | `x' | `1'}.}\label{parse}
\end{minipage}%
\hfill
\begin{minipage}{.48\textwidth}
  \centering
\includegraphics[width=0.9\textwidth]{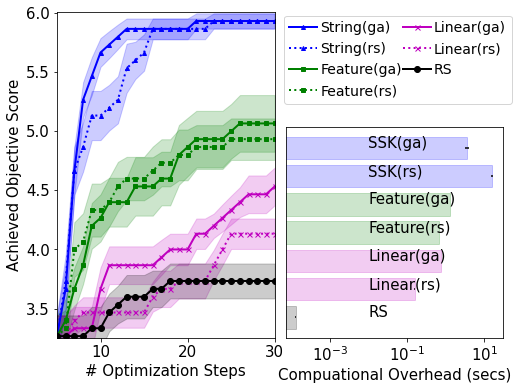}
\caption{Performance and computational overhead when searching for binary strings of length 20 with the most non-overlapping occurrences of "101" (higher is better). }\label{fig::synth}
\end{minipage}
\vskip -0.5cm
\end{figure}

Context free grammars (CFG) are collections of rules able to encode many common syntactical constraints (see Appendix \ref{appendix:cfg} and \cite{hopcroft2001introduction}). While it is difficult to define character-level mutation and crossover operations that maintain grammatical rules over strings of varying length, suitable operations can be defined over parse trees, structures detailing 
the grammatical rules used to make a particular string. Following ideas from grammar-based genetic programming \citep{mckay2010grammar}, mutations randomly replace sub-trees with new trees generated from the same head node, and crossover swaps two sub-trees sharing a head node between two parents (see Figure \ref{parse}). When sampling strings from the grammar to initialize our GA and perform mutations, the simple strategy of building parse trees by recursively choosing random grammar rules produces long and repetitive sequences. We instead employ a sampling strategy that down-weights the probability of selecting a particular rule based on the number of times it has already occurred in the current parse tree branch (Appendix \ref{appendix:GA}).

\section{Experiments}
\label{experiments}
We now evaluate our proposed BO framework on tasks from a range of fields and syntactical constraints. Our code is available at \textit{github.com/henrymoss/BOSS} and is built upon the Emukit Python package \citep{emukit2019}. All results are based on runs across 15 random seeds, showing the mean and a single standard error of the best objective value found as we increase the optimization budget.  The computational overhead of BO (the time spent fitting the GP and maximizing the acquisition function) is presented as average wall-clock times. Although acquisition function calculations could be parallelized across the populations of our GA at each BO step, we use a single-core Intel Xeon 2.30GHz processor to paint a clear picture of computational cost.

\paragraph{Considered BO approaches}

For problems with fixed string-length, we compare our SSK with existing approaches to define GP models over strings. In particular, we apply the squared exponential (SE) kernel \citep{rasmussen2003gaussian} to a bags-of-ngrams  feature representation of the strings. SSKs (feature) representations consider sub-sequences of up to and including five non-contiguous (contiguous) characters, with additional choices demonstrated in Appendix \ref{appendix:synth}. We also provide a linear kernel applied to one-hot encodings of each character, a common approach for BO over categorical spaces. The strategy of sequentially querying random strings is included for all plots and we introduce task-specific baselines alongside their results. After a random initialization of $\min(5,|\Sigma|)$ evaluations, kernel parameters are re-estimated to maximize model likelihood before each BO step.

\begin{table}[t]
\setlength{\tabcolsep}{4pt}
\begin{tabular}{lcr|rcccc}
\multicolumn{3}{c|}{Problem Definition} & \multicolumn{5}{c}{Mean performance with std error (2 s.f.)} \\ \hline
Objective&Space&{\hspace*{-0.25cm}Steps}& SSK (ga) & SSK (rs) & {\hspace*{-0.15cm}}Feature (ga) {\hspace*{-0.15cm}}& {\hspace*{-0.15cm}}Linear (ga){\hspace*{-0.15cm}} & RS \\ \hline
\#  of "101" & $\{0,1\}^{20}$&10 & \textbf{100 (0.0)} & 96 (1.4)  & 97 (2.2)  &58 (3.0)& 58 (2.6)  \\
\#  of "101", 
no overlaps
& $\{0,1\}^{20}$&15 & \textbf{98 (1.4)} & 94 (2.6)  &76 (4.1)  & 64 (2.6) & 60 (3.1)  \\
\cellcolor{red!20}\#  of "10??1" & $\{0,1\}^{20}$&$25$ & \cellcolor{red!20}\textbf{98 (1.6)} &95 (1.6)  & \cellcolor{red!20}64 (2.0) & \cellcolor{red!20}64 (3.3) & 56 (2.0)  \\
\#  of "101" in $1^{\rm st}$ 15 chars
& $\{0,1\}^{30}$& $40$ & \textbf{91 (2.6)} &83 (1.7)  & 67 (3.0)  & 69 (2.6)  & 61 (2.3) \\
\cellcolor{blue!20}\# of "101" 
+ $\mathcal{N}(0,2)$
& $\{0,1\}^{20}$&$25$ & \cellcolor{blue!20}\textbf{98 (2.1)} & 95 (1.4) & \cellcolor{blue!20}51 (3.9)  & \cellcolor{blue!20}40 (4.0)  & 45 (3.8) \\
\cellcolor{yellow!20}\#  of "123" & {\hspace*{-1cm}}$\{0,..,3\}^{30}${\hspace*{-1cm}}&$20$ & \cellcolor{yellow!20}\textbf{81 (2.3)} & \cellcolor{yellow!20}35 (2.8)   &69 (5.4) & 23 (2.0) & 17 (1.5)\\
\cellcolor{yellow!20}\#  of "01??4" & {\hspace*{-1cm}}$\{0,..,4\}^{20}${\hspace*{-1cm}}&$50$ & \cellcolor{yellow!20}\textbf{67 (4.5)}& \cellcolor{yellow!20}38 (2.6)  &35 (4.0)  &33 (3.1)  &29 (2.6) 
\end{tabular}
\vskip 0.2cm
\caption{Optimization of functions counting occurrences of a particular pattern within strings of varying lengths and alphabets ("?" matches any single character). Evaluations are standardized $\in[0,100]$ and higher scores show superior optimization.  Our SSK provides particularly strong performance for complicated patterns (red) or when evaluations are contaminated with Gaussian noise (blue). Our GA acquisition maximizer is especially effective for large alphabets (yellow).}
\label{table:synth}
\vskip -0.5cm
\end{table}

\subsection{Unconstrained Synthetic String Optimization}
\label{unconstrained}
We first investigate a set of synthetic string optimization problems over unconstrained string spaces containing all strings of a particular length built from a specific alphabet. Objective functions are then defined around simple tweaks of counting the occurrence of a particular sub-string. Although these tasks seem simple, we show in Appendix \ref{appendix:synth} that they are more difficult than the synthetic benchmarks used to evaluate standard BO frameworks. The results for seven synthetic string optimization tasks are included in Table \ref{table:synth}, with a deeper analysis of a single task in Figure \ref{fig::synth}. Additional figures for the remaining tasks showing broadly similar behavior are included in the supplement. To disentangle the benefits provided by the SSK and our GA, we consider two acquisition optimizers:  random search across $10,000$ sample strings (denoted \textit{rs} and not to be confused with the random search used to optimize the original objective function) as well as our genetic algorithms (\textit{ga}) limited to  $\leq100$ evolutions of a population of size $100$. The genetic algorithm is at most as computationally expensive (in terms of acquisition function evaluations) as the random search optimizer, but in practice is usually far cheaper due to the GA's early-stopping.

Figure \ref{fig::synth} demonstrates that our approach provides highly efficient global optimization, dramatically out-performing random search and BO with standard kernels. Interestingly, although the majority of our approach's advantage comes from the SSK, our genetic algorithm also contributes significantly to performance, out-performing the random search acquisition function optimizer in terms of both optimization and computational overhead. Although SSKs incur significant BO overheads, they achieve high-precision optimization after far fewer objective queries, meaning a substantial reduction in overall optimization costs for all but the cheapest objective functions. Table \ref{table:synth} shows that our approach provides superior optimization across a range of tasks designed to test our surrogate model's ability to model contextual, non-contiguous and positional information.

\subsection{Locally Constrained Protein Optimization}
\label{genetics}
For our second set of examples, we consider the automatic design of genes that strongly exhibit some particular property. We follow the set-up of \cite{gonzalez2015bayesian}, which optimizes across the space of all the genes encoding a particular protein. Proteins are sequences made from $20$ amino acids, but redundancy in genetic coding means that individual proteins can be represented by many distinct genes, each with differing biological properties. For this experiment, we seek protein representations with minimal \textit{minimum free-folding energy}, a fundamental biological quantity determined by how a protein `folds' in 3-D space. The prediction of the most likely free-folding energy for large sequences remains an important open problem \citep{alquraishi2019alphafold}, whereas calculating the minimal free-folding energy (across all possible folds) is possible for smaller sequences using the ViennaRNA software \citep{lorenz2011viennarna}. We acknowledge that this task may not be  biologically meaningful on its own, however, as free-folding energy is of critical importance to other down-stream genetic prediction tasks, we believe it to be a reasonable proxy for wet-lab-based genetic design loops. This results in a truly challenging black-box string optimization, requiring modeling of positional and frequency information alongside long-range and non-contiguous relationships.

\begin{figure}[t]
\setlength{\tabcolsep}{1pt}
\begin{tabular}{cccc}
  \includegraphics[height=28mm]{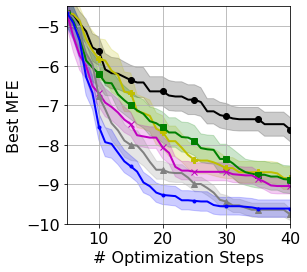} &   \includegraphics[height=28mm]{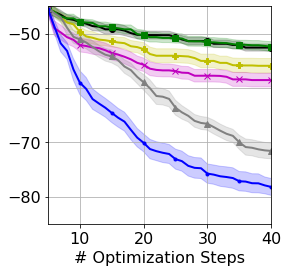} & \includegraphics[height=28mm]{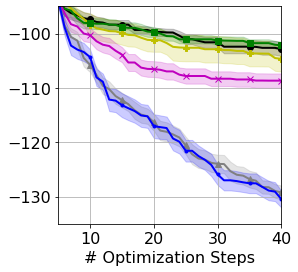} &   \includegraphics[height=28mm]{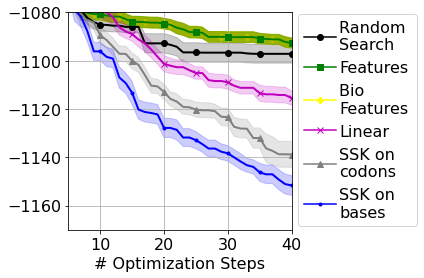}
  \\
$(\ell,m)=(30,1)$ & $(\ell,m)=(186,2)$ & 
$(\ell,m)=(360,8)$ & \hspace{-10mm}$(\ell,m)=(3672,64)$
\end{tabular}
\caption{Finding the representation with minimal \textit{minimum free-folding energy} (MFE) for proteins of length $\ell$. SSKs are applied to codon or base representations split into $m$ or $3m$ parts, respectively.}
\label{genetics_fig}
\vskip -0.5cm
\end{figure}

Each amino acid in a protein sequence can be encoded as one of a small subset of $64$ possible codons, inducing a locally constrained string space of genes, where the set of valid codons depends on the position in the gene (i.e the particular amino acid represented by that position). Alternatively, each codon can be represented as triples of the bases (A,C,T,G), forming another locally constrained string space of three times the length of the codon representation but with a smaller alphabet size of $4$. As well as applying the linear and feature kernels to the base representations, we also consider the domain-specific representation used by \citet{gonzalez2015bayesian} (denoted as Bio-Features) that counts codon frequencies and four specific biologically inspired base-pairs. Figure \ref{genetics_fig} demonstrates the power of our framework across $4$ proteins of varying length. Additional details and wall-clock timing are provided in Appendix \ref{appendix:proteins}. SSKs provide particularly strong optimization for longer proteins, as increasing the length renders the global feature frequencies less informative (with the same representations used for many sequences) and the linear kernel suffers the curse of dimensionality. Note that unlike existing BO frameworks for gene design, our framework explores the large space of all possible genes rather than a fixed small candidate set.

\subsection{Grammar Constrained String Optimization}
\label{subsec:grammareg}
We now consider a string optimization problem under CFG constraints. As these spaces contain strings of variable length and have large alphabets, the linear and feature kernel baselines considered earlier cannot be applied. However, we do consider the VAE-based approaches  of \cite{kusner2017grammar} and \cite{gomez2018automatic} denoted \textit{GVAE} and \textit{CVAE} for a grammar VAE and character VAE, respectively. We replicate the symbolic regression example of \cite{kusner2017grammar}, using their provided VAEs pre-trained for this exact problem. Here, we seek a valid arithmetic expression that best mimics the relationship between a set of inputs and responses, whilst following the syntactical rules of a CFG (Appendix \ref{appendix:cfg}). We investigate both BO and random search in the latent space of the VAEs, with points chosen in the latent space decoded back to strings for objective function evaluations (details in Appendix \ref{appendix:AG}). We sample $15$ strings for initialization of our GPs, which, for the VAE-approaches, are first encoded to the latent space, before being decoded for evaluation. The invalid strings suggested by \textit{CVAE} are assigned large error.

Figure \ref{fig::gram_opt} shows that our approach is able to provide highly efficient BO across a space with complicated syntactical constraints, out-performing the VAE methods which are beaten by even random search (a comparison not made by \cite{kusner2017grammar}). The difference in starting values for the performance curves in  Figure \ref{fig::gram_opt} is due to stochasticity when encoding/decoding; initial strings are rarely decoded back to themselves but instead mapped back to a less diverse set. However, sampling directly in the latent space led to a further decrease in initialization diversity. We stress that \textit{CVAE} and \textit{GVAE} were initially designed as models which, using BO-inspired arguments, could generate new valid strings outside of their training data. Consequently, they have previously been tested only in scenarios with significant evaluation budgets. To our knowledge, we are the first to analyze their performance in the low-resource loops typical in BO.

{
\begin{figure}[t]
\centering
\begin{minipage}[b]{.24\textwidth}
  \centering
  \includegraphics[width=\textwidth]{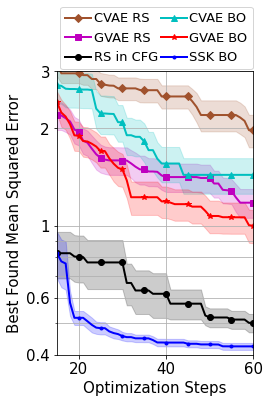}
   \parbox{0.95\textwidth}{\captionof{figure}{\label{fig::gram_opt}Searching for arithmetic expressions satisfying constraints from a CFG (lower is better).}}
\end{minipage}
\hfill
\begin{minipage}[b]{.24\textwidth}
  \centering
  \includegraphics[width=\textwidth]{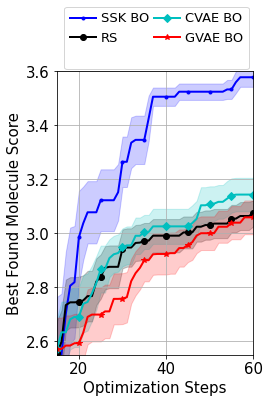}
  \parbox{0.95\textwidth}{\captionof{figure}{\label{fig:SMILES_RESULTS}Searching a candidate set for molecules with desirable properties (higher is better).}}
\end{minipage}
\hfill
\begin{minipage}[b]{.45\textwidth}
  \centering
  \includegraphics[width=\textwidth]{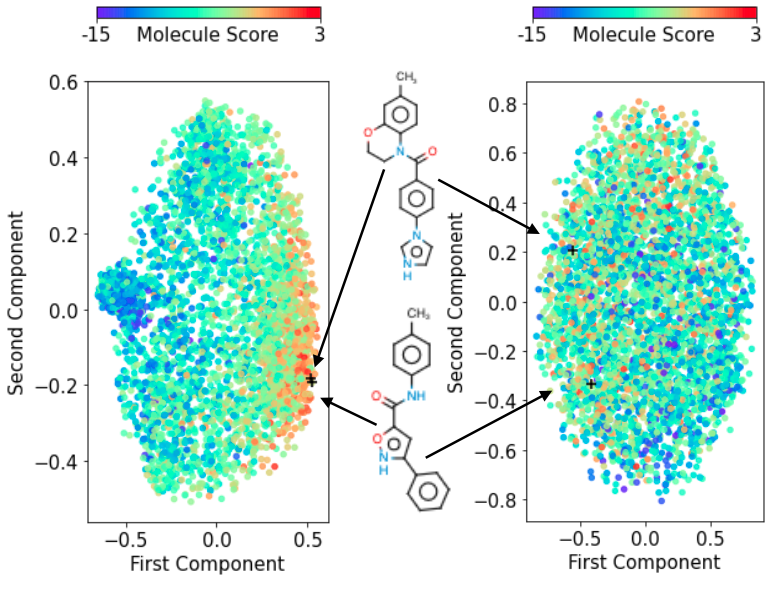} 
  \parbox{0.9\textwidth}{\caption{\label{latents} Top KPCA components for our SSK (left) and an SE kernel in the \textit{GVAE} (right) for SMILES strings. Our SSK has a smoother internal representation, where `close' points are structurally similar.}}
\end{minipage}
\vskip -0.5cm
\end{figure}}

\subsection{Optimization Over a Candidate Set}
\label{samplingeg}

Finally, we return to the task introduced briefly in Figure \ref{workflow} of searching over SMILES strings to find molecules with desirable properties. As the validity of SMILES strings are governed by complex semantic and syntactic rules that can only be partly explained by a context-free grammar \citep{kraev2018grammars}, it is not obvious how to define a GA acquisition function optimizer that can explore the space of all valid molecules. Therefore, we consider an alternative task of seeking high-scoring molecules from within the large collection of $250,000$ candidate molecules used by \cite{kusner2017grammar} to train a \textit{CVAE} and \textit{GVAE}. Once again, we stress that \cite{kusner2017grammar}'s primary motivation is to use a large evaluation budget to generate new molecules outside of the candidate set, whereas we consider the simpler but still realistic task of efficiently exploring within the set's limits. At each BO step, we sample $100$ candidates, querying those that maximize the acquisition function predicted by our SSK as well as by GPs with SE kernels over the VAE latent spaces. Figure \ref{fig:SMILES_RESULTS} shows that only the SSK allows efficient exploration of the candidate SMILES strings. We hypothesize that the VAEs' poor performance may be partly due to  the latent space's dimension which, at $56$, is likely to severely hamper the performance of any BO routine. 

\paragraph{SSK's internal representations}

A common way to investigate the efficacy of VAEs is to examine their latent representations.  However, even if objective evaluations are smooth across this space \citep{kusner2017grammar}, this smoothness cannot be exploited by BO unless successfully encapsulated by the surrogate model. Although GPs have no explicit latent space, they have an intrinsic representation that can be similarly examined to provide visualization of a surrogate model's performance. In particular, we apply kernel principal component analysis (KPCA) \citep{scholkopf1997kernel} to visualize how SMILES strings map into the feature space.  
Figure \ref{latents} shows the first two KPCA components of our SSK and of an SE kernel within the \textit{GVAE}'s latent space (additional visualizations in Appendix \ref{appendix:smiles}). 
Although the latent spaces of the \textit{GVAE} is known to exhibit some smoothness for this SMILES task \citep{kusner2017grammar}, the smoothness is not 
captured by the GP model, in contrast with the SSK.

\section{Discussion}

Departing from fixed-length representations of strings revolutionizes the way in which BO is performed over string spaces. In contrast to VAEs, where models are learned from scratch across thousands of parameters, an SSK's structure is predominantly fixed. By hard-coding prior linguistic intuition about the importance of incorporating non-contiguity, our SSKs have just two easily identifiable kernel parameters governing modeling of a particular objective function. We posit that the additional flexibility of VAEs is not advantageous in BO loops, where there is never enough data to reliably learn flexible models and where calibration is more important than raw predictive strength.

As well as achieving substantially improved optimization, we provide a user-friendly BO building-block that can be naturally inserted into orthogonal developments from the literature, including batch \citep{gonzalez2016batch}, multi-task \citep{swersky2013multi}, multi-fidelity \citep{moss2020mumbo}  and multi-objective \citep{hernandez2016predictive} BO, as well as BO with controllable experimental noise \citep{moss2020bosh} (all with obvious applications within gene and chemical design loops). Moreover, our framework can be extended to other kinds of convolution kernels such as tree \citep{collins2002convolution} and graph kernels \citep{vishwanathan2010graph}. This would allow the optimization of other discrete structures that have previously been modeled through VAEs, including networks \citep{zhang2019d} and  molecular graphs \citep{kajino2018molecular}.


\section*{Broader Impact}

The primary contribution of our work is methodological, providing an efficient and user-friendly framework for optimizing over discrete sequences. As noted in the paper, this is a broad class of problems with a growing interest in the machine learning literature. We hope that our accessible code base will encourage the deployment of our method by practitioners and researchers alike.

We have highlighted two real-world applications by demonstrating efficiency improvements within automatic gene and molecule design loops. Such gains are of considerable interest to biological and chemical research labs. Reducing the wet-lab resources required when searching for chemicals or genes with desirable properties provides not only a substantial environmental and monetary saving, but can even enable new technologies. For example, a fast search over genes is a necessary step in providing custom medical treatments. 

On the other hand, wherever our method can be applied to find structures with beneficial properties, it could similarly be used to find structures with malicious properties. Although the optimization itself is automatic, a human should always has the final say in how a particular optimized structure is to be used. This decision making process should in turn incorporate any ethical frameworks specific to the task at hand.

\section*{Acknowledgments}
The authors are grateful to reviewers, whose comments and advice have improved this paper. The research was supported by EPSRC, the STOR-i Centre for Doctoral Training and a visiting researcher grant from the University of Melbourne.

\def\bibfont{\small}
\bibliography{references}

\newpage
\appendix

\section{Dynamic Programs For SSK Evaluations and Gradients} 
\label{appendix:dynamicprogram}
We now detail recursive calculation strategies for calculating ${k}_n(\textbf{a},\textbf{b})$ and its gradients with $O(nl^3)$ complexity. A recursive strategy is able to efficiently calculate the contributions of particular sub-string, pre-calculating contributions of the smaller sub-strings contained within the target string.

Adapting the recursion and notation of \cite{beck2017learning} to our chosen contribution function, ${k}_n(\textbf{a},\textbf{b})$ can be calculated by following for $i=1,..n$:

\begin{align*}
    \textbf{K}_0'&=\textbf{1}\\
    \textbf{K}'_i &= \textbf{D}^T_{|\textbf{a}|}\textbf{K}''_i\textbf{D}_{|\textbf{b}|}\\
    \textbf{K}_i'' &= \lambda^2_m(\textbf{M}\odot\textbf{K}'_{i-1})\\
    k_i &= \lambda^2_m\sum_{j,k}(\textbf{M}\odot\textbf{K}'_i)_{j,k},
\end{align*}
producing the kernel evaluation ${k}_n(\textbf{a},\textbf{b}) = \sum_{i=1}^nk_i$. Here, $\odot$ is the Hadamard product, $\textbf{M}$ is the $|\textbf{a}|\times|\textbf{b}|$ matrix of character matches between the two strings ($M_{ij}=\mathds{1}_{a_i}(b_j)$), and $\textbf{D}_\ell$ is the $\ell\times\ell$ matrix

\begin{equation}
  \textbf{D}_\ell =\begin{bmatrix}
    0 & 1 & \lambda_g & \cdots & \lambda_g^{\ell-2} \\
    0 & 0 & 1 & \cdots & \lambda_g^{\ell-3} \\
    \vdots & \vdots & \vdots & \ddots & \vdots \\
    0 & 0 & 0 & \cdots & 1\\
    0 & 0 & 0 & \cdots & 0
  \end{bmatrix}.\nonumber
\end{equation}

The gradients of ${k}_n$ with respect to the kernel parameters $\lambda_m$ and $\lambda_g$ can also be  calculated recursively. For the kernel gradients with respect to match decay we calculate

\begin{align*}
    \frac{\partial \textbf{K}_0'}{\partial \lambda_m} &=\textbf{0}\\
    \frac{\partial \textbf{K}_i'}{\partial \lambda_m} &= \textbf{D}^T_{|\textbf{a}|}\frac{\partial \textbf{K}_i''}{\partial \lambda_m} \textbf{D}_{|\textbf{b}|}\\
    \frac{\partial \textbf{K}_i''}{\partial \lambda_m} &= 2\lambda_m(\textbf{M}\odot\textbf{K}'_{i-1}) +
    \lambda_m^2\left(\textbf{M}\odot \frac{\partial \textbf{K}_{i-1}'}{\partial \lambda_m}\right)\\
     \frac{\partial k_i}{\partial \lambda_m} &= \sum_{j,k}\left[ 2\lambda_m(\textbf{M}\odot\textbf{K}'_{ijk}) +
    \lambda_m^2\left(\textbf{M}\odot \frac{\partial \textbf{K}_{ijk}'}{\partial \lambda_m}\right)\right],
\end{align*}
producing the gradient $\frac{\partial {k}_n(\textbf{a},\textbf{b})}{\partial \lambda_m} = \sum_{i=1}^n\frac{\partial k_i}{\partial \lambda_m}$.

Similarly, kernel gradients with respect to gap decay are calculated by
\begin{align*}
    \frac{\partial \textbf{K}_0'}{\partial \lambda_g} &=\textbf{0}\\
    \frac{\partial \textbf{K}_i'}{\partial \lambda_g} &=
    \frac{\partial \textbf{D}^T_{|\textbf{a}|}}{\partial \lambda_g}\textbf{K}_i''\textbf{D}_{|\textbf{b}|}
    +
    \textbf{D}^T_{|\textbf{a}|}\frac{\partial \textbf{K}_i''}{\partial \lambda_g}\textbf{D}_{|\textbf{b}|}
    +
    \textbf{D}^T_{|\textbf{a}|}\textbf{K}_i''\frac{\partial \textbf{D}_{|\textbf{b}|}}{\partial \lambda_g}\\
    \frac{\partial \textbf{K}_i''}{\partial \lambda_g} &= 
    \lambda_m^2\left(\textbf{M}\odot \frac{\partial \textbf{K}_{i-1}'}{\partial \lambda_g}\right)\\
     \frac{\partial k_i}{\partial \lambda_g} &= \lambda_m^2\sum_{j,k}
    \left(\textbf{M}\odot \frac{\partial \textbf{K}_{ijk}'}{\partial \lambda_g}\right), 
\end{align*}
producing the gradient  $\frac{\partial {k}_n(\textbf{a},\textbf{b})}{\partial \lambda_g} = \sum_{i=1}^n\frac{\partial k_i}{\partial \lambda_g}$, where $\frac{\partial \textbf{D}_{\ell}}{\partial \lambda_g}$ is the $\ell\times\ell$ matrix

\begin{equation}
  \frac{\partial \textbf{D}_{\ell}}{\partial \lambda_g} =\begin{bmatrix}
    0 & 0 & 1 & 2\lambda_g & 3\lambda^2_g & \cdots & (\ell-2)\lambda_g^{\ell-3} \\
    0 & 0 & 0 & 1 & 2\lambda_g & \cdots & (\ell - 3)\lambda_g^{\ell-4} \\
    0 & 0 & 0 & 0 & 1 & \cdots & (\ell-4)\lambda_g^{\ell-5} \\
    \vdots & \vdots & \vdots & \vdots & \vdots & \ddots & \vdots \\
    0 & 0 & 0 & 0 & 0 & \cdots 1\\
    0 & 0 & 0 & 0 & 0 & \cdots & 0
  \end{bmatrix}.\nonumber
\end{equation}

\section{Context-free Grammars}
\label{appendix:cfg}

Context-free grammars (CFG) are 4-tuples $G=(V,\Sigma,R,S)$, consisting of:
\begin{itemize}
    \item a set of non-terminal symbols $V$,
    \item a set of terminal symbols $\Sigma$ (also known as an alphabet),
    \item a set of production rules $R$,
    \item a non-terminal starting symbol $S$ from which all strings are generated. 
\end{itemize}

Production rules are simple maps permitting the swapping of non-terminals with other non-terminals or terminals. All strings generated by the CFG can be broken down into a (non-unique) tree of production rules with the non-terminal starting symbol $S$ at its head. These are known as the parse trees and are demonstrated in Figure \ref{parse} in the main paper. 

The CFG for the symbolic regression task of Section \ref{subsec:grammareg} is given by the following rules:

{\centering
\begin{minipage}{.48\textwidth}
\lstinline{S -> S `+' T }\\
\lstinline{S ->  S `*' T }\\
\lstinline{S -> S `/' T }\\
\lstinline{S ->  T}\\
\lstinline{T -> `(' S `)'}\\
\lstinline{T ->  `sin(' S `)'}\\
\lstinline{T ->  `exp(' S `)'}\\
\lstinline{T -> `x'}\\
\lstinline{T ->  `1' }\\
\lstinline{T ->  `2' }\\
\lstinline{T -> `3'},\\
\end{minipage}}

where $V=\{S,T\}$ and $\Sigma=\{+,*,/,x,1,2,3\}$. Although each individual production rule is a simple replacement operation, the combination of many such rules can specific a string space with complex syntactical constraints. For example, these $11$ rules are able to specify that the string `(sin(2*x)+3(x*(2+exp(x))))+1/2' is valid but that `(sin(2*x)+3(x*(2+exp(x)))+1/2' (with invalid bracket closing) is not.

\textbf{Sampling from the CFG}. One of the advantages of CFGs is that it is easy (and cheap) to generate large collections of valid strings by recursively sampling production rules. However, when sampling strings from the grammar, we found this simple sampling strategy to produce long and repetitive strings. For our BO applications, where sample diversity is key, we instead employed a sampling strategy that down-weights the probability of selecting a particular rule based on the number of times it has already occurred in the parse tree. In particular, the probability of applying a particular rule to a non-terminal is proportional to $c^n$, where $n$ is the number of occurrences of that rule in the current branch and $c$ is a discount factor (set to $0.1$ in all our experiments). The construction of this sampler ensures that a wide range of production rules are used when generating from the CFG.

\section{Genetic Algorithms}
\label{appendix:GA}

We now provide implementation details for our GA acquisition function optimizers. During each GA step, populations are refined through stochastic biologically-inspired operations, providing a population achieving (on average) higher scores.   The GA begins with a randomly sampled population and ends once the best string in the population stops improving between iterations (Algorithm \ref{alg}). The $N$ strings of the $i+1^{th}$ population are perturbations of the $i^{th}$ population. To evolve a population (Algorithm \ref{EVOLVE}), a \textit{tournament} process first selects $n$ candidate strings (with replacement) attaining the highest evaluations across random sub-samples of a proportion $p_t$ of the current population. To create the next population, these candidate strings undergo stochastic perturbations: a \textit{mutation} operation producing a new offspring string from a single parent, and a \textit{crossover} operation combining attributes of two parent strings to produce two new offspring. These operations occur with probability $p_c$ and $p_m$ respectively, which, alongside $p_t$, control the level of  diversity maintained across populations. To highlight the robustness of our genetic algorithm acquisition optimizer, we do not tune the evolution parameters to each task, using populations of $100$ candidate strings and $(p_t,p_c,p_m)=(0.5,0.75,0.1)$ for all our experiments. The exact crossover and mutation operators chosen to traverse string spaces under different syntactical constraints are discussed in the main paper.

\begin{algorithm}[t]
\begin{algorithmic}[1]
\Function{\textbf{GA}}{$p_t$, $p_c$, $p_m$, $N$}
\State $n\leftarrow0$
\State Sample $N$ strings for initial population  $P_0$ 
\State Evaluate acquisition function  $A_0\leftarrow\alpha(P_0)$ 
\State Store current best value $\alpha_{best}\leftarrow\max(A_0)$
\While {$\alpha_{best}=\max(A_n)$}
    \State Begin new iteration $n\leftarrow n+1$
    \State Evolve population $P_n\leftarrow \textbf{EVOLVE}(P_{n-1},p_t,p_c,p_m)$
    \State Evaluate acquisition function  $A_n\leftarrow\alpha(P_n)$ 
    \State Store current best value $\alpha_{best}\leftarrow \max(\max(A_{n-1}),\alpha_{best})$
\EndWhile
\State \Return String achieving score $\alpha_{best}$
\EndFunction
\end{algorithmic}
\caption{\label{alg}Genetic Algorithms for Acquisition Function Maximization}
\end{algorithm}

\begin{algorithm}[t]
\begin{algorithmic}[1]
\Function{\textbf{EVOLVE}}{$P$,$p_t$, $p_c$, $p_m$}
\State Initialize new population $P_{new}\leftarrow\emptyset$
\While {$|P_{new}|<|P|$}
    \State Collect a candidate string $s_1\leftarrow\textbf{TOURNAMENT}(P,p_t)$
    \State Sample $r\sim U[0,1]$
    \If {$r<p_c$} 
        \State Sample another candidate string $s_2\leftarrow\textbf{TOURNAMENT}(P,p_t)$
        \State Perform crossover $s_1,s_2 \leftarrow \textbf{CROSSOVER}(s_1,s_2)$
        \State Sample $r_1,r_2\sim U[0,1]$
        \If {$r_1<p_m$}
            \State Perform mutation $s_1 \leftarrow \textbf{MUTATION}(s_1)$
        \EndIf
        \If {$r_2<p_m$}
            \State Perform mutation $s_2 \leftarrow \textbf{MUTATION}(s_2)$
        \EndIf
        \State Add two strings to new population $P_{new}\leftarrow P_{new}\bigcup\{s_1,s_2\}$
    \Else
        \State Sample $r\sim U[0,1]$
        \If {$r_1<p_m$}
            \State Perform mutation $s_1 \leftarrow \textbf{MUTATION}(s_1)$
        \EndIf
        \State Add string to new population $P_{new}\leftarrow P_{new}\bigcup\{s_1\}$
    \EndIf
\EndWhile
\State \Return New population $P_{new}$
\EndFunction
\end{algorithmic}
\caption{\label{EVOLVE}Evolution of Genetic Algorithm Populations}
\end{algorithm}

\newpage
\section{Synthetic String Optimization Experiments}
\label{appendix:synth}

\begin{figure}[t]
\centering
\begin{minipage}{.48\textwidth}
  \centering
\includegraphics[width=\textwidth]{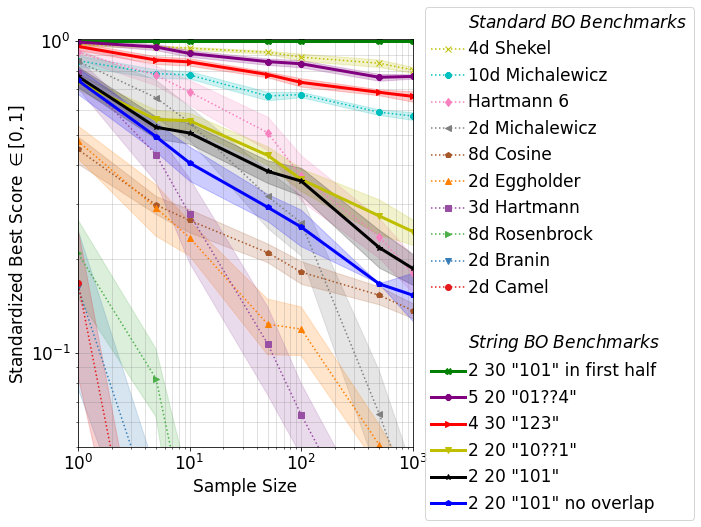}
\caption{\label{benchmarks}Comparing random search across standard BO benchmarks (faint) and our synthetic string experiments (bold). For the string tasks, the legend $A L S$ denoted the task with an alphabet of size $A$, strings of length $L$ and counting the occurrences of the pattern $S$.}
\end{minipage}
\hfill
\begin{minipage}{.48\textwidth}
  \centering
\includegraphics[width=\textwidth]{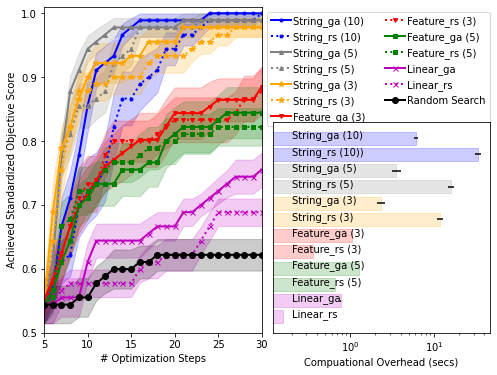}
\caption{\label{results_1}Optimizing the number of non-overlapping occurrences of "101" in a string of length 20 and alphabet ["0","1"]}
\end{minipage}
\end{figure}

Although seemingly simple tasks, our synthetic string optimization tasks of Section \ref{unconstrained} are deceptively challenging, as only a very small proportion of valid strings produce high scores. In fact, these tasks are considerably more challenging than the common benchmarks used to test standard BO frameworks. Figure  \ref{benchmarks}, shows the performance attained by random search over our synthetic string tasks and standard benchmarks \footnote{\textit{https://www.sfu.ca/~ssurjano/index.html}}. All objective functions are standardized ($\in[0,1]$) and we run $1000$ optimization steps, plotting the mean and standard error across $25$ replications. We see that our easiest synthetic string optimization tasks are among the hardest of the standard benchmark problems to solve with random search, and we expect this to hold similarly for BO.

\begin{figure}[t]
\centering
\begin{minipage}{.48\textwidth}
  \centering
\includegraphics[width=\textwidth]{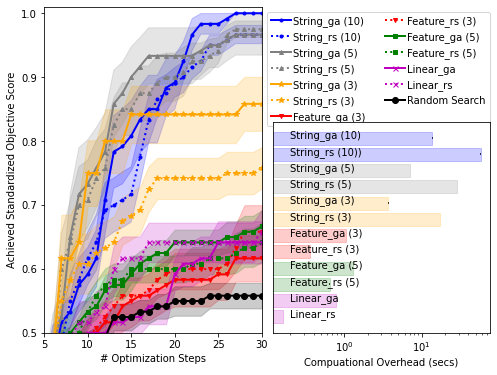}
\caption{\label{results_2}Optimizing the number of occurrences of "10??1" in a string of length 20 and alphabet ["0","1"]}
\end{minipage}%
\hfill
\begin{minipage}{.48\textwidth}
  \centering
\includegraphics[width=\textwidth]{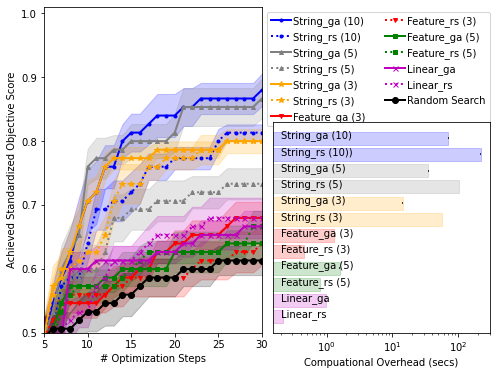}
\caption{\label{results_3}Optimizing the number of occurrences of "101" in the first half of a string of length 30 and alphabet ["0","1"].}
\end{minipage}
\end{figure}

\begin{figure}[H]
\centering
\begin{minipage}{.48\textwidth}
  \centering
\includegraphics[width=\textwidth]{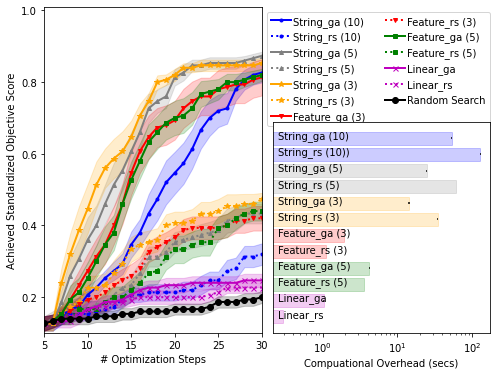}
\caption{\label{results_4}Optimizing the number of occurrences of "123" of a string with length 30 and an alphabet of ["0","1","2","3"]. }
\end{minipage}%
\hfill
\begin{minipage}{.48\textwidth}
  \centering
\includegraphics[width=\textwidth]{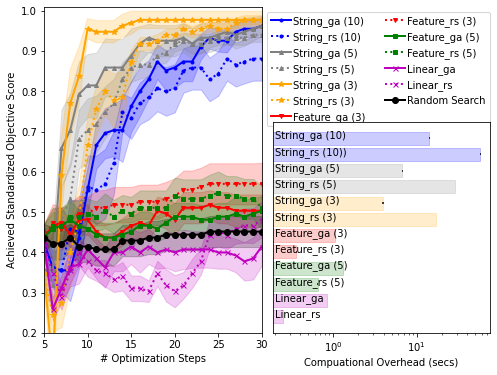}
\caption{\label{results_5}Optimizing the number of occurrences of "101" with observations contaminated by Gaussian noise (with a variance of 2) of a binary string of length 20. }
\end{minipage}
\end{figure}

\begin{figure}[H]
\centering
\begin{minipage}{.48\textwidth}
  \centering
\includegraphics[width=\textwidth]{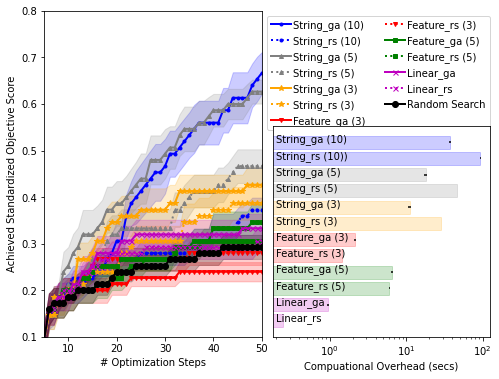}
\caption{\label{results_6}Optimizing the number of occurrences of "01??4" in a string of length 20 and alphabet ["0","1","2,"3","4"]}
\end{minipage}%
\hfill
\begin{minipage}{.48\textwidth}
  \centering
\includegraphics[width=\textwidth]{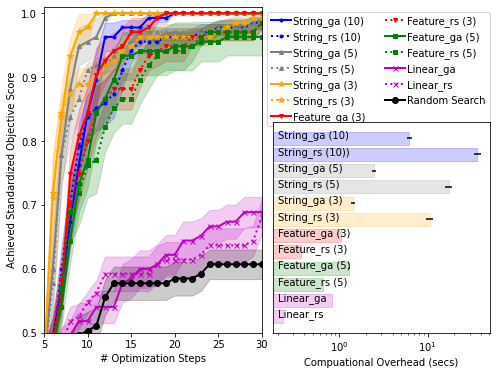}
\caption{\label{results_7}Optimizing the number of occurrences of "101" in a string of length 20 and alphabet ["0","1"]}
\end{minipage}%
\end{figure}

We now provide comprehensive experimental results across the synthetic string optimization tasks. In Figures \ref{results_1},\ref{results_2},\ref{results_3},\ref{results_4},\ref{results_5},\ref{results_6} and \ref{results_7}, we show the performance and computational overhead of our string kernels, extending the analysis from the main paper to include a variety of sub-sequence lengths considered by the string and feature-based kernels. We see that the string kernels always provide superior optimization over existing kernels, with the string kernel based on sub-sequences of maximum length $5$ consistently among the best. The string kernel is particularly effective for the most complicated objective functions (Figures \ref{results_2} and \ref{results_6}) and when observations are contaminated by observation noise (Figure \ref{results_5}). For problems with larger alphabets (and so significantly larger search spaces), our genetic algorithm acquisition optimizer dramatically outperforms a larger budget random search optimizer (Figure \ref{results_4} and \ref{results_6}).


\newpage
\section{Protein Optimization}
\label{appendix:proteins}
We now provide additional details for our four protein optimization experiments, each targeting one of the following proteins.
\begin{enumerate}
    \item Cystic fibrosis transmembrane conductance regulator:
\begin{verbnobox}[\fontsize{8pt}{8pt}\selectfont]
        TIKENIFGVS.
\end{verbnobox}
    \item Invertebrate iridescent virus 6 (IIV-6) (Chilo iridescent virus):
\begin{verbnobox}[\fontsize{8pt}{8pt}\selectfont]
        MTSRGHLRRAPCCYAFKSATSHQRTRTSLCLASPPAPHCLLLYSHRCLTYFTVDYELSFFCL.
\end{verbnobox}
    \item Anaphase-promoting complex subunit 15B:
\begin{verbnobox}[\fontsize{8pt}{8pt}\selectfont]
        MSTLFPSLLPQVTDSLWFNLDRPCVDENELQQQEQQHQAWLLSIAEKDSSLVPIGKPASEPY
        DEEEEEDDEDDEDSEEDSEDDEDMQDMDEMNDYNESPDDGEIEADMEGAEQDQDQWMI.
\end{verbnobox}
    \item Tyrosine-protein kinase abl-1:  
\begin{verbnobox}[\fontsize{8pt}{8pt}\selectfont]
        MGHSHSTGKEINDNELFTCEDPVFDQPVASPKSEISSKLAEEIERSKSPLILEVSPRTPDSV
        QMFRPTFDTFRPPNSDSSTFRGSQSREDLVACSSMNSVNNVHDMNTVSSSSSSSAPLFVALY
        DFHGVGEEQLSLRKGDQVRILGYNKNNEWCEARLYSTRKNDASNQRRLGEIGWVPSNFIAPY
        NSLDKYTWYHGKISRSDSEAILGSGITGSFLVRESETSIGQYTISVRHDGRVFHYRINVDNT
        EKMFITQEVKFRTLGELVHHHSVHADGLICLLMYPASKKDKGRGLFSLSPNAPDEWELDRSE
        IIMHNKLGGGQYGDVYEGYWKRHDCTIAVKALKEDAMPLHEFLAEAAIMKDLHHKNLVRLLG
        VCTHEAPFYIITEFMCNGNLLEYLRRTDKSLLPPIILVQMASQIASGMSYLEARHFIHRDLA
        ARNCLVSEHNIVKIADFGLARFMKEDTYTAHAGAKFPIKWTAPEGLAFNTFSSKSDVWAFGV
        LLWEIATYGMAPYPGVELSNVYGLLENGFRMDGPQGCPPSVYRLMLQCWNWSPSDRPRFRDI
        HFNLENLISSNSLNDEVQKQLKKNNDKKLESDKRRSNVRERSDSKSRHSSHHDRDRDRESLH
        SRNSNPEIPNRSFIRTDDSVSFFNPSTTSKVTSFRAQGPPFPPPPQQNTKPKLLKSVLNSNA
        RHASEEFERNEQDDVVPLAEKNVRKAVTRLGGTMPKGQRIDAYLDSMRRVDSWKESTDADNE
        GAGSSSLSRTVSNDSLDTLPLPDSMNSSTYVKMHPASGENVFLRQIRSKLKKRSETPELDHI
        DSDTADETTKSEKSPFGSLNKSSIKYPIKNAPEFSENHSRVSPVPVPPSRNASVSVRPDSKA
        EDSSDETTKDVGMWGPKHAVTRKIEIVKNDSYPNVEGELKAKIRNLRHVPKEESNTSSQEDL
        PLDATDNTNDSIIVIPRDEKAKVRQLVTQKVSPLQHHRPFSLQCPNNSTSSAISHSEHADSS
        ETSSLSGVYEERMKPELPRKRSNGDTKVVPVTWIINGEKEPNGMARTKSLRDITSKFEQLGT
        ASTIESKIEEAVPYREHALEKKGTSKRFSMLEGSNELKHVVPPRKNRNQDESGSIDEEPVSK
        DMIVSLLKVIQKEFVNLFNLASSEITDEKLQQFVIMADNVQKLHSTCSVYAEQISPHSKFRF
        KELLSQLEIYNRQIKFSHNPRAKPVDDKLKMAFQDCFDQIMRLVDR.
\end{verbnobox}
\end{enumerate}

As each amino acid in these protein sequences can be represented as one of a set of possible codons (triples of bases), the string spaces for these problems are incredibly large, with each space containing $5.53\mathrm{e}{+4}$, 
$9.48\mathrm{e}{+33}$, $4.81\mathrm{e}{+49}$ and $1.22\mathrm{e}{+614}$ unique strings, respectively. The permitted mappings from amino acids to valid codons are as follows:

{\centering
\begin{minipage}{.48\textwidth}
\lstinline{F -> ttt|ttc}\\
\lstinline{L -> tta|ttg|ctt|ctc|cta, ctg}\\
\lstinline{S -> tct|tcc|tca|tcg|agt|agc}\\
\lstinline{Y -> tat|tac}\\
\lstinline{C -> tgt|tgc}\\
\lstinline{W -> tgg}\\
\lstinline{P -> cct|ccc|cca|ccg}\\
\lstinline{H -> cat|cac}\\
\lstinline{Q -> caa|cag}\\
\lstinline{R -> cgt|cgc|cga|cgg|aga|agg}\\
\lstinline{I -> att|atc|ata}\\
\lstinline{M -> atg}\\
\lstinline{T -> act|acc|aca|acg}\\
\lstinline{N -> aat|aac}\\
\lstinline{K -> aaa|aag}\\
\lstinline{V -> gtt|gtc|gta|gtg}\\
\lstinline{A -> gct|gcc|gca|gcg}\\
\lstinline{D -> gat|gac}\\
\lstinline{E -> gaa|gag}\\
\lstinline{G -> ggt|ggc|gga|ggg}.\\
\end{minipage}}

\begin{figure}[t]
\begin{tabular}{cc}
  \includegraphics[width=60mm]{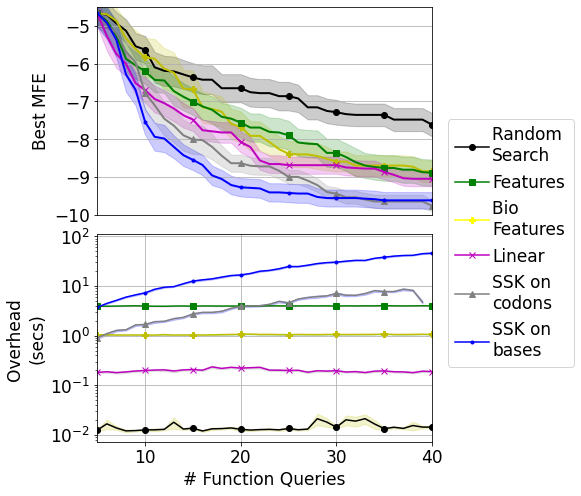} &   \includegraphics[width=60mm]{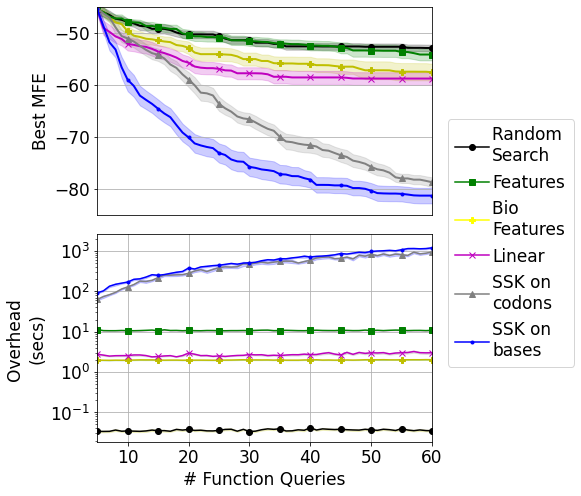} \\
(a) $(\ell,m)=(30,1)$ & (b) $(\ell,m)=(186,2)$ \\[6pt]
 \includegraphics[width=60mm]{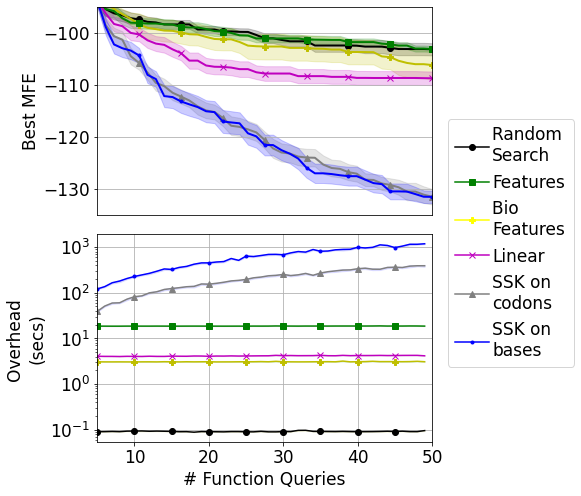} &   \includegraphics[width=60mm]{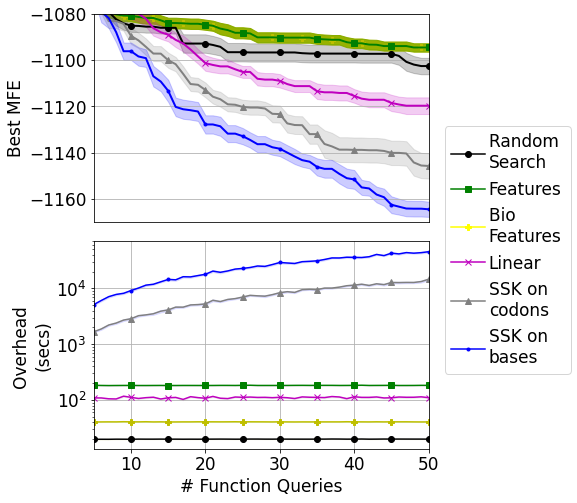} \\
(c) $(\ell,m)=(360,8)$ & (d) $(\ell,m)=(3672,64)$ \\[6pt]
\end{tabular}
\caption{\label{extra_protein}Optimization performance and computational overhead when finding the representation with minimal \textit{minimum free-folding energy} (MFE) of a protein of length $\ell$. SSKs are applied to codon or base representations split into $m$ or $3m$ parts, respectively.}
\end{figure}

Figure \ref{extra_protein} extends the analysis of our protein optimization tasks to include the computational overheads incurred by each each BO routine (as measured on a single processor). The high evaluation costs of our SSK means that its overhead is substantially greater than the other approaches. However, in real gene design loops, this additional computational cost (hours) is negligible compared to the cost and time saved in wet-lab experiments (days). Moreover, the acquisition function calculations can be trivially parallelized across up to $100$ cores (the size of the populations used in the GA acquisition function optimizer) as well as across the $m$ partial SSK calculations. If GPUs are available, these can also be used to efficiently calculate SSKs \citep{beck2017learning}.

\section{BO in a VAE's Latent Space}
\label{appendix:AG}
To perform BO in the latent space of a VAE, we follow the set-up of \cite{kusner2017grammar}, fitting a GP with an SE kernel and using a multi-start gradient descent acquisition function optimizer. We tried SE kernels with both individual and tied length scales across latent dimensions, however, this did not have a significant effect on performance, possibly due to difficulties in estimating many kernel parameters in these low-data BO problems. In order to perform BO, a compact area of the latent space must be chosen for the search space. Unfortunately, \cite{kusner2017grammar} do not provide details about how this should be determined. We chose the space containing the most central $75\%$  of representations from the set of strings used to train the VAE ($100,000$ arithmetic expressions). We also tried using the space containing all representations from the training data, however, this led to a drop in optimization performance, likely due to less reliable encoding/decoding learned by the VAE in these more sparsely supported parts of the latent space.

\section{Visualizing BO Surrogate Models}
\label{appendix:smiles}
\begin{figure}[t]
\begin{tabular}{cc}
  \includegraphics[width=60mm]{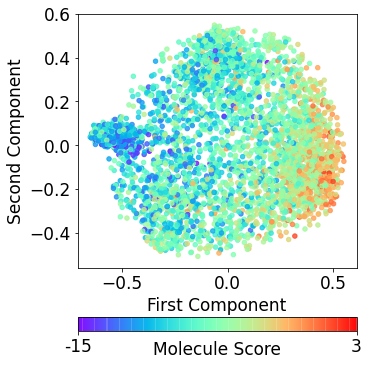} &   \includegraphics[width=60mm]{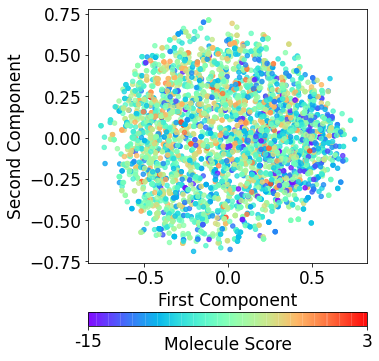} \\
(a) SSK on raw SMILES strings. & (b) SE kernel in the \textit{CVAE} latent space. \\[6pt]
 \includegraphics[width=60mm]{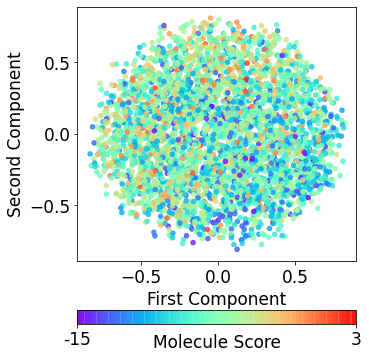} &   \includegraphics[width=60mm]{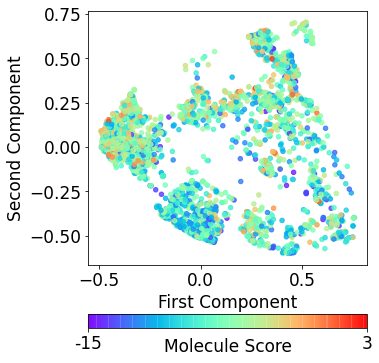} \\
(c) SE kernel in the \textit{GVAE} latent space. & (d) SSK with poor choices of kernel parameters. \\[6pt]
\end{tabular}
\caption{Top two KPCA components visualizing the intrinsic representations of the surrogate models used to predict molecule scores from SMILES strings. Aside from (d), kernel parameters are tuned to maximize GP likelihood over $10$ evaluated molecules.}
\label{KPCA}
\end{figure}

In Section \ref{samplingeg}, we present a kernel principal component analysis (KPCA) visualization of the feature space induced by our SSK. We now extend this analysis to include the VAE competitors. In particular, we perform KPCA on the SE kernel used to define a surrogate model over each VAE's latent representations (Figure \ref{KPCA}). All figures show the representations of the same sampled $4,000$ SMILES strings, color-coded to represent their molecule scores (a linear combination of their water-octanol partition coefficient, ring-size and synthetic accessibility). We see that the GP with an SSK produces a significantly smoother KPCA space that the GPs fit in VAE latent space, with the \textit{CVAE} showing slightly more structure than the \textit{GVAE}. This ranking matches the relative performance of the BO routines based on these surrogate models (Figure \ref{fig:SMILES_RESULTS}). So although the latent spaces of these VAE have been shown to exhibit some smoothness \citep{kusner2017grammar}, this is not captured by the GP model. Figure \ref{fig:SMILES_RESULTS}.d visualizes the intrinsic representation of an SSK when kernel parameters are purposely chosen to provide a bad fit. We choose very low $\lambda_m$ and high $\lambda_g$ to heavily penalize the long contiguous sub-sequences we know to be informative for this task. The stark difference in smoothness between the visualizations of the tuned and badly-tuned SSKs demonstrates their flexibility as well as the importance of using a representation supervised to the the specific objective function of interest.

\end{document}